\documentclass[10pt,twocolumn,letterpaper]{article}

\usepackage{wacv}              % To produce the CAMERA-READY version
\usepackage{graphicx}
\usepackage{amsmath}
\usepackage{amssymb}
\usepackage{booktabs}

\usepackage[pagebackref,breaklinks,colorlinks]{hyperref}
\usepackage[accsupp]{axessibility} 

\usepackage[capitalize]{cleveref}
\crefname{section}{Sec.}{Secs.}
\Crefname{section}{Section}{Sections}
\Crefname{table}{Table}{Tables}
\crefname{table}{Tab.}{Tabs.}

\def\wacvPaperID{326} % *** Enter the WACV Paper ID here
\def\confName{WACV}
\def\confYear{2025}

\begin{document}

%%%%%%%%% TITLE - PLEASE UPDATE
\title{ARF-Plus: Controlling Perceptual Factors in Artistic Radiance Fields for 3D Scene Stylization}

\author{
Wenzhao Li\textsuperscript{\rm 1}, 
    Tianhao Wu\textsuperscript{\rm 1}, 
    Fangcheng Zhong\textsuperscript{\rm 1}, 
    Cengiz Oztireli\textsuperscript{\rm 1}\\
\textsuperscript{\rm 1}University of Cambridge\\
{\tt\small wl301@cantab.ac.uk, tw554@cam.ac.uk, fz261@cam.ac.uk, aco41@cam.ac.uk }
% For a paper whose authors are all at the same institution,
% omit the following lines up until the closing ``}''.
% Additional authors and addresses can be added with ``\and'',
% just like the second author.
% To save space, use either the email address or home page, not both
}
\maketitle

% \author{First Author\\
% Institution1\\
% Institution1 address\\
% {\tt\small firstauthor@i1.org}

% For a paper whose authors are all at the same institution,
% omit the following lines up until the closing ``}''.
% Additional authors and addresses can be added with ``\and'',
% just like the second author.
% To save space, use either the email address or home page, not both
% \and
% Second Author\\
% Institution2\\
% First line of institution2 address\\
% {\tt\small secondauthor@i2.org}
% }

% \author {
%     % Authors
%     Wenzhao Li\textsuperscript{\rm 1},
%     Tianhao Wu\textsuperscript{\rm 1},
%     Fangcheng Zhong\textsuperscript{\rm 1}
%     Cengiz Oztireli\textsuperscript{\rm 1}
% }
% \Institution {
%     % Affiliations
%     \textsuperscript{\rm 1}University of Cambridge\\
%     % \textsuperscript{\rm 2}Google Research\\
%     wl301@cam.ac.uk, tw554@cam.ac.uk, fz261@cam.ac.uk, aco41@cam.ac.uk 
% }
% \maketitle

%%%%%%%%% ABSTRACT
\begin{abstract}
    The radiance fields style transfer is an emerging field that has recently gained popularity as a means of 3D scene stylization, thanks to the outstanding performance of neural radiance fields in 3D reconstruction and view synthesis. We highlight a research gap in radiance fields style transfer, the need for sufficient perceptual controllability, motivated by the existing concept in the 2D image style transfer. In this paper, we present ARF-Plus, a unique 3D neural style transfer framework offering manageable control over perceptual factors, to systematically explore the perceptual controllability in 3D scene stylization. Four distinct types of controls - color preservation control, (style pattern) scale control, spatial (selective stylization area) control, and depth enhancement control - come with our proposed novel loss functions and strategies, seamlessly integrated into this framework. This unlocks a realm of limitless possibilities, allowing customized modifications of stylization effects and flexible merging of the strengths of different styles, ultimately enabling the creation of novel and eye-catching stylistic effects on 3D scenes.
\end{abstract}

%%%%%%%%% BODY TEXT
\section{Introduction}
\label{sec:intro}

\begin{figure}[t]
\setlength{\belowcaptionskip}{-0.6cm}
\centering
\includegraphics[width=1\linewidth]{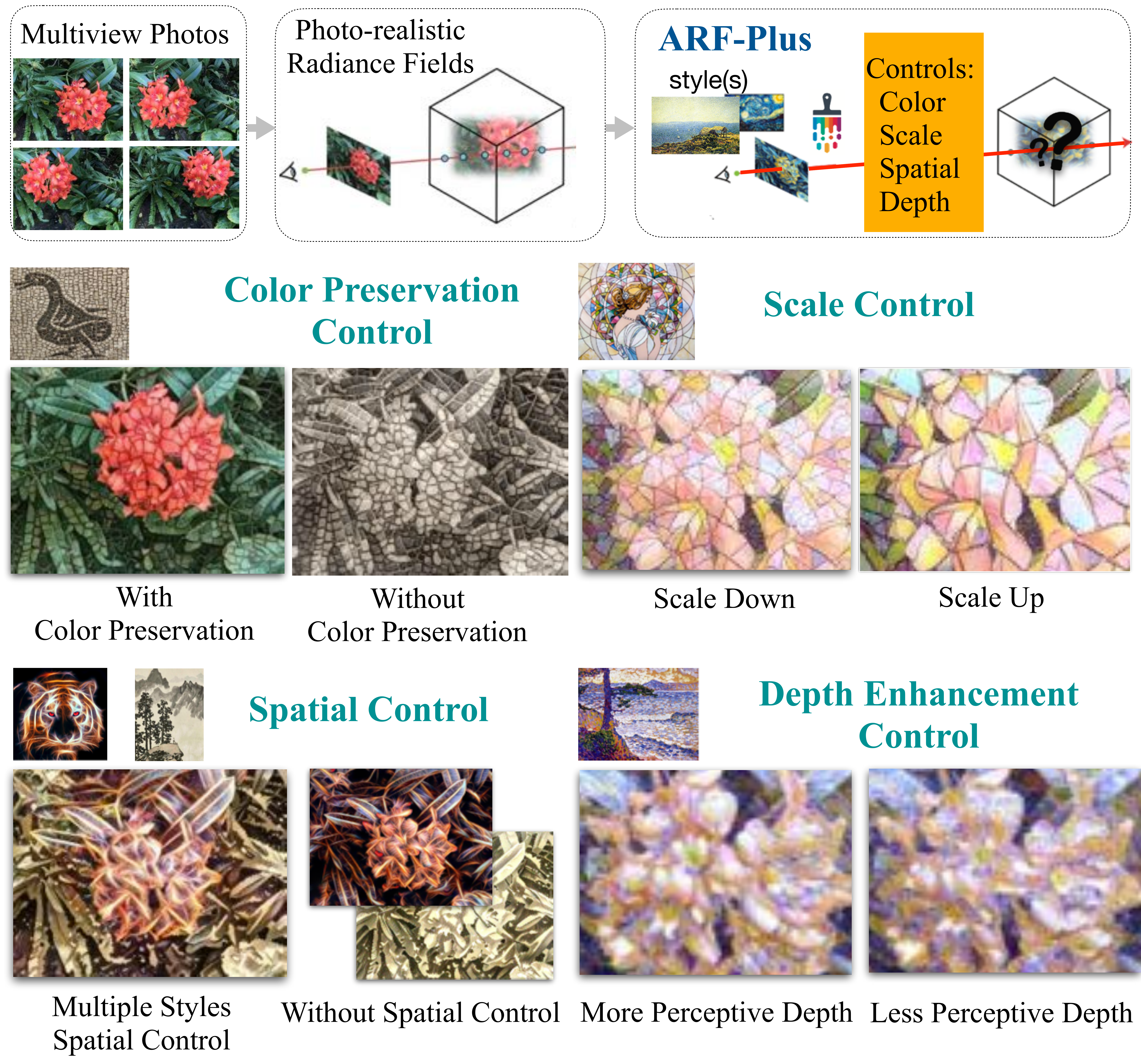}
  \caption{An overview of ARF-Plus framework. It facilitates the control of various perceptual factors and permits the use of multiple style images as input, resulting in perceptual flexibility in 3D scene stylization.
}
  \label{fig:framework}
\end{figure}

The neural style transfer technique \cite{gatys2016image} unravels the mysteries of the artist. With the aid of this technology, a photograph may look to have been produced by a well-known artist. 3D scene style transfer further extends it to the spatial-temporal domain, offering enhanced artistic transformations and immersive aesthetic experiences. Nonetheless, 3D scene style transfer is not an easy task due to the domain gap between 2D and 3D. Traditional 3D scene stylization techniques ~\cite{huang2021learning, mu20223d, yin20213dstylenet, hollein2022stylemesh}  are often based on mesh or point-cloud representations. 
The limited reconstruction accuracy and discrete geometric representation in these models constrain the stylization performance, resulting in observable artifacts in challenging real-world circumstances~\cite{zhang2022arf}. The Neural Radiance Fields (NeRF)~\cite{mildenhall2021nerf} represents the optics of the 3D scene without explicit geometry which is inherently discontinuous. With its superior performance in 3D reconstruction and view synthesis, many researchers have attempted to perform 3D stylization based on radiance fields~\cite{chiang2022stylizing, huang2021learning, huang2022stylizednerf, zhang2022arf}. However, they have focused on optimizing a unified style transfer across the whole scene, neglecting the control of perceptual factors that influence visual perception.

Color preservation control~\cite{gatys2016preserving}, stroke size control~\cite{jing2018stroke, reimann2022controlling}, spatial control~\cite{yu2019method, gatys2017controlling}, and other types of perceptual controls have been widely applied in the image domain style transfer. The research gap in the radiance fields style transfer hinders the diversity of choices and potential applications in 3D scene style transfer. Users might want to apply an abstract style to a scene's background, keeping the main objects realistic. They could also wish to add unique effects to specific objects for a cartoon-like look. Some may prefer retaining the scene's original colors while applying patterns from a chosen style. Additionally, users might adjust patterns or brushstroke sizes for aesthetic preferences. All these diverse 3D radiance fields style transfer requirements and numerous promising application scenarios are yet to be fully realized.

Hence, exploring perceptual control in 3D stylization on radiance fields and developing algorithms that offer perceptual-feature-aware controls hold promising potential for meeting application-level demands. Our main contributions are listed as follows:
\begin{itemize}
  \item We introduce ARF-Plus, a framework incorporated with novel methodologies that can \emph{simultaneously} manipulate various perceptual factors in 3D stylization. 
  % The combination use of our control methods offers more flexibility and creates novel stylization outcomes.
  To the best of our knowledge, we are the \emph{first} to explore the control of various perceptual factors in 3D stylization based on radiance fields.
  
  \item We demonstrate the effectiveness of ARF-Plus in controlling the aforementioned perceptual factors on 3D stylization through a series of qualitative and quantitative evaluations.

\end{itemize}  

\section{Background and Related Work}
\label{sec:background}

3D style transfer entails applying a reference style's visual characteristics to a target 3D scene while maintaining the 3D structure itself. Due to the exceptional 3D reconstruction capabilities of radiance fields, radiance fields style transfer has emerged as a promising approach to enhance the outcomes of 3D style transfer. Recent studies~\cite{chiang2022stylizing, fan2022unified, huang2022stylizednerf, nguyen2022snerf, zhang2022arf, chen2022upst} have shown that style features can be transferred from a given style to an entire real 3D scene represented by radiance fields.

Zhang Kai et al.~\cite{zhang2022arf} present the Artistic Radiance Fields (ARF) approach, which is highly effective at capturing style characteristics while preserving multi-view consistency. They present and implement a nearest neighbour-based feature matching (NNFM) loss, which can produce a consistent multi-view visual experience and achieve state-of-the-art visual quality. Additionally, their approach is independent of the radiance fields representation, making it applicable to a wider range of systems, including Plenoxels~\cite{yu2021plenoctrees}, NeRF~\cite{mildenhall2021nerf}, TensoRF~\cite{chen2022tensorf}, and others. We choose ARF as the basis of our architecture for two primary reasons. Firstly, ARF is currently one of the best-performing state-of-the-art methods. It captures style details while maintaining multi-view consistency and produces more convincing outcomes than other state-of-the-art techniques such as Huang et al.~\cite{huang2022stylizednerf} and Chiang et al.~\cite{chiang2022stylizing}. Secondly, the ARF method is versatile and adaptable to various radiance fields representations. As a result, our ARF-Plus can benefit from ARF's exceptional performance and inherit ARF's broad applicability.

Based on our best knowledge, the concept of perceptual control has not been well utilized in the radiance fields style transfer. NeRF-Art~\cite{wang2023nerf} achieves text-driven neural radiance fields stylization, by addressing the mapping between a text prompt and a particular given style or aesthetic pattern. However, its matched stylized result with a given text prompt is still a uniformly stylized scene. Ref-NPR~\cite{zhang2023ref} proposes a ray registration process, in order to maintain cross-view consistency and establish semantic correspondence in transferring style across the entire stylized scene. While Ref-NPR mentioned the concept of ``controllable", its notion of controllability pertains to semantic correspondence and geometrically consistent stylizations. This differs from the user-level controllability discussed in this paper, where we aim to offer enhanced flexibility, enabling individuals to effectively customize controlled stylization. StyleRF \cite{liu2023stylerf} introduces the zero-shot manner in 3D Nerf stylization which can generalize non-existent stylization via multi-style interpolation. Although it mentions the spatial composition results of two different styles, the primary function of the 3D consistent segmentation masks is to define spatial boundaries. These masks guide the zero-shot model in producing new styles for corresponding geometric regions without additional training. This differs from our approach to spatial control. Additionally, StyleRF focuses more on generating entirely new styles that don't exist, while our paper primarily discusses achieving a more customized effect by controlling different aspects or making fine-tuned adjustments and combinations of known styles.

% \textcolor{red}{StyleRF \cite{liu2023stylerf} introduces the zero-shot manner in 3D Nerf stylization which can generalize non-existent stylization via multi-style interpolation. Although it mentions the spatial composition results of two different styles, the primary function of the 3D consistent segmentation masks is to define spatial boundaries. These masks guide the zero-shot model in selecting different styles for corresponding geometric regions without additional training. This differs from our approach to spatial control. Additionally, StyleRF focuses more on generating entirely new styles that don't exist, while our paper primarily discusses achieving a more customized effect by controlling different aspects or making fine-tuned adjustments and combinations of known styles.} 

Nevertheless, similar to the demand for customized and diverse 2D image style transfer \cite{wu2022completeness, wu2020efanet, song2019etnet, liu2021adaattn, hu2020aesthetic}, there exists a substantial market for radiance field style transfer with personalized and varied characteristics. Our primary objective is to fill this research gap, and the subsequent sections will present our initial endeavours and discoveries.

\section{Methodology}
\label{sec:methodology}
As illustrated in Fig.\ref{fig:framework}, we begin by reconstructing photo-realistic radiance fields from multiple photographs to represent a 3D scene. Subsequently, we fine-tune the radiance fields with given style image(s), formulating the style transfer as an optimization problem. A variety of our proposed loss functions and gradient updating strategies are employed based on the selected perceptual controls. After the stylization process is completed, we achieve the stylized 3D scene radiance fields and can generate free-viewpoint and consistent stylized renderings from it.

\subsection{Color Preservation Control}
% The general radiance fields style transfer algorithms convert the photo-realistic radiance fields, which depict a real 3D scene, to match the style of a specified style image. Once the stylization process is complete, 
% In general radiance fields style transfer, the color distribution of the style reference is also transferred. However, in some cases, one may prefer to retain the colors of the photo-realistic 3D scene and only learn the painting patterns, while ensuring that stylization does not result in unnatural colors (e.g., the sky turning green or the grass becoming red).

In radiance fields style transfer, color distribution is typically transferred from the style reference, but in some cases, there is a preference for retaining realistic 3D scene colors while solely learning painting patterns to avoid unnatural color outcomes (e.g. a green sky).

Our approach builds upon the luminance-only algorithm ~\cite{gatys2016preserving} but integrates novel perspectives to form a novel unified loss function. The work ~\cite{gatys2016preserving} in 2D image domain is based on the utilization of the $YIQ$ color space, which effectively separates luminance in $\mathit{Y}$, and color information in $\mathit{I}$ and $\mathit{Q}$. They implement style transfer to generate an output image in $\mathit{Y}$ first and then a post-processing technique ``channel-concatenation'' - appending $\mathit{I}$ and $\mathit{Q}$ to $\mathit{Y}$ - is applied to obtain a color image. However, radiance fields models consist of neural networks (e.g. NeRF \cite{mildenhall2021nerf}) or voxel grids with spherical harmonics (e.g. Plenoxel \cite{yu2021plenoctrees}), which are not explicitly and solely represented by color channels concatenation.

As shown in Eq. \ref{eq:color-control-lum}, we propose a novel loss equation. We extract the luminance channel in the rendered view $\hat{\mathbf{X}^{Y}}$ and the luminance channel of the style image $\mathbf{X}_{\text{style}}^{Y}$ for style loss $\mathcal{L}_{\text{style}}$ calculation, while keeping the RGB channels for the content loss $\mathcal{L}_{\text{content}}$ calculation.

\setlength{\abovedisplayskip}{5pt}
\begin{equation}\label{eq:color-control-lum}
% \setlength{\abovedisplayskip}{-5pt}
% \vspace{-5pt}
% \setlength{\belowdisplayskip}{3pt}
\begin{split}
 \scalebox{1}{$\mathcal{L}_{\text{total}}
  = \ $} & \scalebox{1}{$ \alpha \cdot \mathcal{L}_{\text{style}}(F_{\ell_{s}}(\hat{\mathbf{X}^{Y}}), F_{l_{s}}(\mathbf{X}_{\text{style}}^{Y})) $}
  \\ & 
  \scalebox{1}{$ + \  \beta \cdot \mathcal{L}_{\text{content}}(F_{\ell_{c}}(\hat{\mathbf{X}}), F_{\ell_{c}}(\mathbf{X}_{\text{content}})) + \gamma \cdot \mathcal{L}_{\text{tv}}$}
 \end{split}
\end{equation}

\noindent The style loss $\mathcal{L}_{\text{style}}$ enables the radiance fields to focus on the luminance variations present in the style image, without explicitly learning its color. The content loss $\mathcal{L}_{\text{content}}$ quantifies the RGB color differences between the rendered view $\hat{\mathbf{X}}$ and the photo-realistic ground truth $\mathbf{X}_{\text{content}}$ at the level of content features. $\mathcal{L}_{\text{tv}}$ is the total variation regularizer. $F_{\ell_{c}}$ and $ F_{\ell_{s}}$ stand for the representation of content features and style features, respectively. $\alpha$, $\beta$ and $\gamma$ are scalar weights.

\subsection{Scale Control}
% \FZ{better define strokes and patterns first.}
% Stroke sizes and pattern shapes in drawing refer to two fundamental aspects of mark-making and line creation within an artistic composition. Stroke sizes pertain to the thickness or width of lines, brushstrokes, or marks used in a drawing, as shown in Fig. \ref{fig:scale_control_both_new} (a). 
% % \ww{\st{Varying stroke sizes can create contrast, depth, and visual interest within the artwork. Thicker strokes might be used to define outlines or add emphasis, while finer strokes can be employed for details or delicate elements. The manipulation of stroke sizes contributes to the overall texture and dynamics of the drawing.}} 
% Pattern shapes involve the specific forms, designs, or arrangements of strokes or marks that are repeated or used consistently within the artwork, as shown in Fig.\ref{fig:scale_control_both_new} (b). 

Artists have their own way of arranging strokes of different sizes and shape patterns, which reflects unique artistic ``styles". For instance, Pablo Picasso favored simple shapes and a limited color palette when painting, whereas Claude Monet employed small, playful brush strokes to create his impressionist works. In certain instances, the stroke and pattern sizes from the original style image may not suit the 3D scene after style transfer. This is due to the differences (e.g. in size) between the style image content and the 3D scene. Adjusting the pattern scale for the 3D scene is vital for enhancing aesthetic appeal.
% scale control to adjust the scaling of coarse and
% fine detail patterns for the best aesthetic appearance.

% For instance, Pablo Picasso favored simple shapes and a limited color palette when painting objects, people, and landscapes, whereas Claude Monet employed small, playful brush strokes to create his impressionist works. In certain instances, the stroke and pattern sizes from the original style image may not suit the 3D scene after style transfer. This is due to the differences (e.g. in size) between the style image content and the 3D scene. Adjusting the pattern scale to align with the 3D scene becomes a crucial consideration.

% It is worth noting that, in some cases, the stroke and pattern sizes of the original style image may not be appropriate for the 3D scene after the style transfer has been performed, due to the differences (e.g. in size) between the depicted content in the style image and the 3D scene. Therefore, adjusting the pattern scale to match the 3D scene becomes an issue that requires attention.

% it becomes necessary to adjust the pattern scale to match the 3D scene. 

% % Adjusting the pattern scale to match the 3D scene becomes an issue that requires attention.

% \fz{Therefore, it becomes necessary to align the pattern scale with the 3D scene.}

\subsubsection{Single Style Scale Control} 
% We implement a modification to xx's work in the field of image style transfer, enabling precise scale control within the 3D domain.

% We apply a modification of Jing Y. et al.~\cite{jing2018stroke} on stroke-controllable image style transfer. 

We implement a modification to 2D image stroke-controllable style transfer ~\cite{jing2018stroke}, enabling precise scale control on the 3D domain. we take into account the two factors mentioned in their work that influence the scale of style patterns: the receptive field and the size of the style image. 

% We propose scale control methods in our ARF-Plus framework, which can achieve flexible and continuous style pattern scaling. In order to facilitate scale management for either a single style or multiple blended styles, we propose two versions of the style loss function. In addition to successfully scaling the overall style pattern, it is worth noting that our techniques enable selective scaling of fine or coarse patterns from the given style(s). 

% Similar to ARF~\cite{zhang2022arf}, our ARF-Plus utilizes the pre-trained VGG-16~\cite{simonyan2014very} to calculate both style and content loss to enable style transfer. In VGG-16, the receptive field size undergoes a substantial increase as the network progresses to deeper layers. Additionally, shallow layers do better in extracting fine details (local features) from the style image, whereas deeper layers perform better at capturing wider and coarse patterns. The size of the style image also contributes to the scaling effect of the style, as the VGG features vary with the scale of the input image. With a rise in the size of the style image, the style pattern typically grows larger in stylized output.
\setlength{\abovedisplayskip}{5pt}
\begin{equation} \label{eq:scale control single loss style}
% \setlength{\abovedisplayskip}{-3pt}
% \vspace{-3pt}
% \scalebox{0.8}{$x = \frac{-b \pm \sqrt{b^2 - 4ac}}{2a}$}
% \scalebox{2}{$x = \frac{-b \pm \sqrt{b^2 - 4ac}}{2a}$}
\scalebox{1}{$
   \mathcal{L}^{\mathcal{T}_{S}}_{\text{style}}=\sum_{l \in \left\{l_1, ..., l_5\right\}} w_{l} \cdot \mathcal{L}_{\text{style}}\left(\mathcal{F}_l\left(\mathcal{R}\left(\mathbf{X}_{\text{style}}, \mathcal{T}^l_{S}\right)\right),\mathcal{F}_l\left(\hat{\mathbf{X}}\right)\right)
$}
\end{equation}

As shown in Eq. \ref{eq:scale control single loss style}, $\mathcal{F}_l$ represents the feature map at $l$ layer blocks, where $l$ can be one of the five VGG-16 layer blocks $\left\{l_1, l_2, ... , l_5\right\}$. We adjust the range of the receptive field by altering the selection of $l$. $\mathcal{R}$ denotes the resize function for the style image. According to the specified scale $\mathcal{T}^l_{S}$ for each layer block, $\mathcal{R}$ resizes the style image for each layer block for feature extraction. The style losses from different layer blocks are subsequently combined together to form the total style loss $\mathcal{L}^{\mathcal{T}_{S}}_{\text{style}}$, with $w_{l}$ controlling the corresponding weight. By having the flexibility to adjust both the size of the style image and the range of the receptive field, we are able to attain continuous control over the scale. 

\begin{equation} \label{eq:scale control single loss total}
% \setlength{\abovedisplayskip}{-5pt}
% \vspace{-2pt}
\scalebox{1}{$
    \mathcal{L}_{\text{total}}=\alpha \cdot \mathcal{L}^{\mathcal{T}_{S}}_{\text{style}} + \beta \cdot \mathcal{L}_{\text{content}}(F_{\ell_{c}}(\hat{\mathbf{X}}), F_{\ell_{c}}(\mathbf{X}_{\text{content}})) + \gamma \cdot \mathcal{L}_{\text{tv}}
$}
\end{equation}

Multiple settings might be purposefully employed to highlight either the delicate patterns or the harsh outlines within a single style image. Giving various weights to the losses of shallow and deep layers will achieve this. The limitless adjustment of scaling settings also allows for the creation of innovative aesthetic patterns. The total style loss $\mathcal{L}^{\mathcal{T}_{S}}_{\text{style}}$ is then added to the style transfer loss function to provide scale control in 3D scene stylization, as shown in Eq. \ref{eq:scale control single loss total}. 

There are three key aspects where our work differs from that of Jing Y. et al.~\cite{jing2018stroke}: 1) In contrast to 2D image style transfer, we use the improved style transfer loss function for 3D radiance fields style transfer, leading to a different overall model architecture. 2) In order to ensure consistency across multi-view scenes, we utilize NNFM loss~\cite{zhang2022arf} rather than the Gram matrix to calculate the style loss. 3) We optimized their continuous size control strategy. In addition to assigning different weights to the output feature maps, we allow alternative style image size $\mathcal{T}^l_{S}$ for each VGG layer in order to produce more adjustable results. In other words, it is possible to have different $\mathcal{T}^l_{S}$ for each layer in our method.

% \WW{This sounds more like hyperparameter tuning?}\WL{The point here is for each layer, we may have different $\mathcal{T}^l_{S}$. However in~\cite{jing2018stroke}, $\mathcal{T}^l_{S}$ for each layer is same. } \WW{I see, please just state that clearly} \WLL{Fixed. If it is still unclear, could you please assist by directly revising my description? thanks}

\subsubsection{Multiple Style Scale Control} 
There may arise a need to blend multiple styles. For instance, it may be desirable to incorporate the fine brushstrokes of one painting with the bigger shapes apparent in another artwork and blend them seamlessly into the 3D scene. It allows for fusing the benefit of two styles in this way. As shown in Eq. \ref{eq: scale-control-blend-style-loss} and \ref{eq: scale-control-blend-total-loss}, we apply our proposed method to situations that encompass a blend of various styles.

\begin{equation}\label{eq: scale-control-blend-style-loss}
% \setlength{\abovedisplayskip}{-13pt}
% \vspace{-4pt}
\scalebox{1}{$
    \mathcal{L}^{\mathcal{T}_{M}}_{\text{style}} = \sum_{i}\lambda_{i}\mathcal{L}^{\mathcal{T}_{S}}_{\text{style}_{i}}
$}
\end{equation}

\begin{equation}\label{eq: scale-control-blend-total-loss}
\scalebox{1}{$
    \mathcal{L}_{\text{total}}=\alpha \cdot \mathcal{L}^{\mathcal{T}_{M}}_{\text{style}} + \beta \cdot \mathcal{L}_{\text{content}}(F_{\ell_{c}}(\hat{\mathbf{X}}), F_{\ell_{c}}(\mathbf{X}_{\text{content}})) + \gamma \cdot \mathcal{L}_{\text{tv}}
$}
\end{equation}

\noindent The scalar weight $\lambda_{i}$ regulates the influence of each individual style image $\mathbf{X}_{\text{style}_{i}}$. The combination loss $\mathcal{L}^{\mathcal{T}_{M}}_{\text{style}}$ serves as the style loss term in the style transfer loss $ \mathcal{L}_{\text{total}}$.

\subsection{Spatial Control}

General 3D style transfer approaches typically apply the style globally to the entire scene, producing a uniform stylization throughout the whole scene. Yet, in specific scenarios, it's crucial to stylize only a specific area or object in a scene. For example, creating an immersive virtual world by giving the background a fantasy watercolor effect while leaving the people or main subjects unchanged.

% there are circumstances where it is necessary to only stylize a specific area or object within a scene. For instance, one may desire to create an immersive virtual world experience by turning the background environment into a fantasy watercolor painting effect without affecting the people or main subjects of a scene. 
% Another example is creating an augmented reality effect by turning the primary objects of a scene into cartoon-like, abstract figures while keeping the background intact. 

% The diverse scenarios mentioned above all require the implementation of spatial control to achieve. We propose a simple but efficient spatial control method in our ARF-Plus framework, which has the ability to selectively impose certain aesthetic styles on particular areas or objects within a 3D scene.

\subsubsection{Single Style Spatial Control}
Our proposed spatial control is integrated with the deferred back-propagation technique in ARF~\cite{zhang2022arf}. During the deferred back-propagation process, the loss and gradient of the predicted view are calculated with auto-differentiation disabled. Each 2D predicted view is associated with a cached-gradients map of the same size.
We design a binary spatial mask $\mathbf{T}^{r}$ to the cached-gradients map $\hat{\mathbf{X}}_{\text{grad}}$ to achieve spatial control - only the gradients corresponding to the masked region are retained, as shown in Eq.  \ref{eq: spatial-control-single}, where $\circ$ denotes element-wise multiplication. 

Each element in $\mathbf{T}^{r}$ takes either a value of 0 or 1. The regions in the mask with a value of 0 correspond to the portions of the predicted view that are not subject to stylization, while the regions with a value of 1 reflect the areas of the view that need to be stylized. As a result, only the gradients corresponding to the chosen regions are used in the computation throughout the optimization step where the cached gradients are back-propagated.

\begin{equation}\label{eq: spatial-control-single}
\scalebox{1}{$
    \hat{\mathbf{X}}_{\text{grad}} = \mathbf{T}^{r} \circ \hat{\mathbf{X}}_{\text{grad}}
$}
\end{equation} The source of the spatial mask is not limited and can be manually annotated. In this paper, we employ semantic segmentation maps $\mathbf{T}_{\text{semantic}}^{r}$ to determine the regions of specific semantic objects within the scene.

% we propose two approaches for obtaining spatial masks: 1. Utilizing the binary segmented depth map $\mathbf{T}_{depth}^{r}$ to identify important object regions. This is based on the assumption that, within the user's field of view, the main objects that grab their interest typically have less depth than the surroundings. Since the reconstructed photo-realistic radiance fields (used as the stylization input) contains depth information for each view, the depth map can be directly generated. 2. Employing semantic segmentation maps $\mathbf{T}_{semantic}^{r}$ to determine the regions of specific semantic objects within the scene. This method is particularly suitable for cases with significant variations in viewpoints. 

\subsubsection{Multiple Style Spatial Control}
To accomplish the spatial control of multiple styles, we introduce a novel strategy called ``Combined Cached-gradients Map'', which still utilizes the concept of the cached-gradients map in ARF~\cite{zhang2022arf}. Assuming the scene contains $r$ areas that need individually corresponding to a unique style image $\mathbf{X}_{\text{style}}^{r}$. As shown in Eq. \ref{eq: spatial-control-multi-loss-function} and \ref{eq: spatial-control-multiple-styles}, for each area $\mathbf{T}^{r} \circ \hat{\mathbf{X}}$ and its related style $\mathbf{X}_{\text{style}}^{r}$, we initially determine the corresponding cached gradients maps $\hat{\mathbf{X}}_{\text{grad}}^{r}$ according to the style transfer loss $\mathcal{L}_{\text{total}}^{r}$ with auto-differentiation disabled. $\hat{\mathbf{X}}_{\text{grad}}^{r}$ are then added to generate the final cached gradients maps $\hat{\mathbf{X}}_{\text{grad}}$ applied in deferred back-propagation. During the optimization process, each selected region is updated according to its target style.

\begin{equation}\label{eq: spatial-control-multi-loss-function}
\begin{split}
\scalebox{1}{$
\mathcal{L}_{\text{total}}^{r}  = \  $}
& \scalebox{1}{$\alpha \cdot \mathcal{L}_{\text{style}}(F_{\ell_{s}}(\mathbf{T}^{r} \circ \hat{\mathbf{X}}), F_{l_{s}}(\mathbf{X}_{\text{style}}^{r})) + \gamma \cdot \mathcal{L}_{\text{tv}} $} \\
 & \scalebox{1}{$ + \ \beta \cdot \mathcal{L}_{\text{content}}(F_{\ell_{c}}(\mathbf{T}^{r} \circ \hat{\mathbf{X}}), F_{\ell_{c}}(\mathbf{T}^{r} \circ \mathbf{X}_{\text{content}})) 
  $}
\end{split}
\end{equation}

\begin{equation}\label{eq: spatial-control-multiple-styles}
\scalebox{1}{$
\hat{\mathbf{X}}_{\text{grad}} = \sum_{r}^{}\mathbf{T}^{r} \circ \hat{\mathbf{X}}_{\text{grad}}^{r}
$}
\end{equation}

\subsection{Depth Enhancement Control}
% While the style transfer technique yields impressive effects by evenly applying style patterns across the entire scene, there are instances where it inadvertently compromises the scene's depth perception. 

% According to some research findings~\cite{liu2017depth, jing2019neural}, stylized results are especially unsatisfactory for situations with a large range of depths. This happens as a result of the uncontrollable alteration of the content layout, which obscures the line separating the foreground from the background and the boundaries between various objects.

Research findings~\cite{liu2017depth, jing2019neural} suggest that stylized outcomes are notably dissatisfactory in scenarios with extensive depth variations. This occurs due to uncontrolled alterations in content layout, obscuring distinctions between foreground and background, as well as object boundaries.

% In order to better maintain the original layout and relative depth relationships during the stylization process of 3D scenes, w
 % Our depth-aware control technique is a variant of Liu et al~\cite{liu2017depth}'s depth-aware approach for 2D image style transfer. 

% Our technique is a variation of Liu et al.~\cite{liu2017depth}, extending the depth-aware approach from 2D image style transfer to 3D scenes. 

Our technique extends 2D image depth-aware style transfer ~\cite{liu2017depth} to 3D scenes by incorporating a depth consistent loss via an off-the-shelf monocular depth estimator. Liu et al.~\cite{liu2017depth} note that the depth map serves as an effective representation of the spatial distribution within an image, and proposes to preserve the depth map of the content during stylization. It is important to note that, to achieve smooth gradient optimization, ARF~\cite{zhang2022arf} fixes the density component in the radiance fields during stylization. Thus, the actual depth information within the radiance fields remains unchanged, while only the appearance of the 3D scene is altered. However, as the style transfer is done in a 2D per-view way without any knowledge of the scene depth or geometry, object boundaries can be easily blurred or lost during stylization. Hence, we need to rely on a pre-trained depth estimation network $\phi_{1}$~\cite{ranftl2020towards} to approximate the perceived depth of the current stylized appearance, as the depth information remains fixed in the radiance field. Our proposed strategy can assist ARF-Plus in promptly understanding the changes in stylization's appearance and whether they would result in harmful alterations to the visually perceived depth.

To preserve depth-aware information, we take the outputs of the pre-trained depth estimation network $\phi_{1}$ and compute the (squared, normalized) Euclidean distances as the $\mathcal{L}_{\text{depth}}^{\phi_{1}}$. The photo-realistic view ${\mathbf{X}_{\text{content}}}$ and stylized view $\mathbf{\hat{X}}$ have the same size $H \times W \times C$. 

\begin{equation}\label{eq: depth-control-depth-loss}
\scalebox{1}{$
\mathcal{L}_{\text{depth}}^{\phi_{1}} = \frac{\left \|\phi_{1} (\mathbf{\hat{X}}) -\phi_{1} (\mathbf{X}_{\text{content}}) \right \|_{2}}{H \times W \times C} 
$} 
\end{equation}
\setlength{\belowdisplayskip}{5pt}

\noindent As shown in Eq. \ref{eq: depth-control-total-loss}, the depth loss $\mathcal{L}_{\text{depth}}^{\phi_{1}}$ is incorporated into the style transfer loss $\mathcal{L}_{\text{total}}$ , where $\delta$ is a scalar weight. $\mathcal{L}_{\text{depth}}^{\phi_{1}}$ serves as an additional aspect to guide and influence the stylization based on depth-aware information.

% \begin{equation}
% \mathcal{L}_{depth}^{\phi_{1}} = \omega_{\phi_{1}} \frac{\left \|\phi_{1} (\mathbf{\hat{X}}) -\phi_{1} (\mathbf{X}_{content}) \right \|_{2}}{C\times H \times W} 
% \end{equation}
\vspace{-5pt}
\begin{equation} \label{eq: depth-control-total-loss}
\scalebox{1}{$
    \mathcal{L}_{\text{total}}=\alpha \cdot \mathcal{L}_{\text{style}} + \beta \cdot \mathcal{L}_{\text{content}} + \delta \cdot \mathcal{L}_{\text{depth}}^{\phi_{1}}  + \gamma \cdot \mathcal{L}_{\text{tv}} 
$}    
\end{equation}
\vspace{-8pt}

%-------------------------------------------------------------------------
\section{Experiments and Evaluation}
\label{sec:experiments}

\subsection{Experiments Design}
% \subsubsection{Baseline}
\noindent \textbf{Baseline}\quad To the best of our knowledge, we are the first to systematically address the issue of enhancing perceptual flexibility controls for radiance fields style transfer. As such, there are no similar works in the current state-of-the-art that can be compared to ours. Since our proposed ARF-Plus framework is based on the ARF \cite{zhang2022arf}, we choose it as the baseline for our experimental evaluation.

% \subsubsection{Datasets}
\noindent \textbf{Datasets}\quad To guarantee comparability with the baseline and achieve more accurate differentiation, we utilise the same datasets as those employed in ARF \cite{zhang2022arf}.\\\textbf{a) Artworks Dataset} \cite{zhang2022arf}: This dataset is made up of a variety of images which act as given styles. It is used to assess the effectiveness of our proposed ARF-Plus framework in learning and managing multiple styles while adhering to specific perceptual factor controls. \\\textbf{b) LLFF Dataset }\cite{ mildenhall2019local}: This dataset contains real-world scenes that are photographed from forward angles. It is utilized to create photo-realistic radiance fields, representing the 3D scenes prior to stylization. \\\textbf{c) Real 360 Degree Scenes Dataset} \cite{knapitsch2017tanks}: This dataset includes real, unbounded, 360-degree large outdoor scenes with complex geometric layouts and camera trajectories. The scene data is also employed to train photo-realistic radiance fields before stylization.

\noindent \textbf{Training Settings}\quad Identical to the baseline, in ARF-Plus, we configure the stylization optimization to run for 10 epochs, with a learning rate that exponentially decreases from 1e-1 to 1e-2. For single-style input, the training time of our ARF-Plus is comparable to that of baseline, as it does not entail additional training tasks during the stylization optimization process, which typically takes around 20 minutes on a single GPU.

% \subsubsection{Implementation Details} 

\noindent \textbf{Implementation Details}\quad Our ARF-Plus framework, as previously mentioned, comprises reconstructing photo-realistic radiance fields from numerous pictures and then stylizing the resulting reconstruction with perceptual factor choices. 1) Regarding the first step of reconstructing a photo-realistic radiance field, our work aligns with the baseline approach in primarily relying on Plenoxels \cite{yu2021plenoctrees} to achieve. Plenoxels is renowned for its fast reconstruction and rendering speed. All training settings of our method in this process remain consistent with the baseline. 2) Although ARF \cite{zhang2022arf}'s core algorithm achieves acceptable style transfer in both style pattern and color, its authors introduce a recoloring strategy in the stylization process to enhance visual quality. In ARF-Plus, we continue to employ the recoloring method for scale control and depth enhancement control in single-style cases. However, we choose to disable the recoloring process in scenarios where color preservation control, spatial control, and multiple input styles are involved. In those instances, it is inappropriate to follow the recoloring process in ARF which uses the colors of a given style image to recolor the entire scene. To ensure a fair comparison, when our ARF-Plus deactivates the recolor process, the baseline ARF also disables this step accordingly. Please refer to the Appendix for additional technical details.

\subsection{Color Preservation Control}

% In the context of color preservation control, our goal is to ensure that the colors of the stylized scene remain consistent with those of the photo-realistic radiance fields. 

\begin{figure}[tb]
\centering
\setlength{\belowcaptionskip}{-0.4cm}
\includegraphics[width=1\linewidth]{  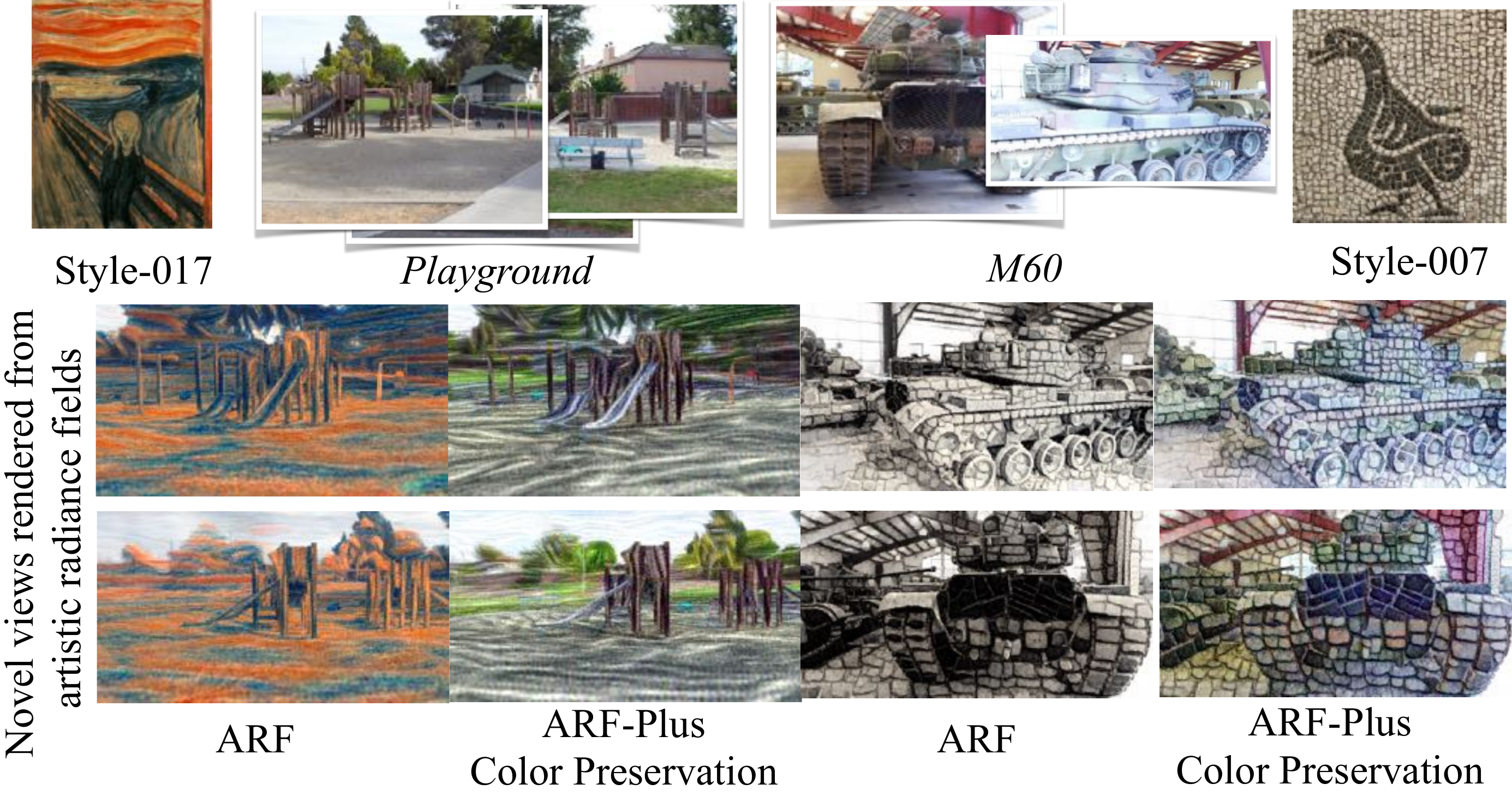}
% \caption{Qualitative results of color preserve control: style transfer on the real-world forward-facing scenes \textit{Leaves}, \textit{Flower}, \textit{Trex} and \textit{Horns}. Our proposed method, ARF-Plus with Color Preserve Control, successfully preserves the original colors of the 3D scene while effectively learning the patterns from the style image. Compared to the color histogram matching approach, the luminance-only algorithm yields better results.
% Supplementary videos are provided [\href{https://drive.google.com/drive/folders/15zoJA8WtGAZs-JQGzIWGEDmcPFfPa_Xa?usp=sharing}{\textcolor{red}{click}}].}

% Supplementary videos are provided [\href{https://drive.google.com/drive/folders/15zoJA8WtGAZs-JQGzIWGEDmcPFfPa_Xa?usp=sharing}{\textcolor{red}{click}}].

  \caption{Qualitative results of color preservation. Please refer to supplementary materials for better visualization.}
  \label{fig:color_control_360}
\end{figure}

Fig. \ref{fig:color_control_360} demonstrates visual comparisons between methods applied to real-world scenes. As illustrated, our proposed color preservation control method effectively preserves the original scene's color while successfully learning the desired style. Moreover, the colors and brightness distribution of the stylized outputs achieved through our method closely resemble those of the photo-realistic view. Please refer to the Appendix for additional qualitative and quantitative evaluations.

\subsection{Scale Control}
% \wll{Quantitative evaluation for scale control is not conducted, and specific reasons are explained in the Appendix. Further experimental outcomes regarding the study of influential factors are also available in the Appendix.} 
We demonstrate the effects of scale control on both single and multiple styles. Please refer to the Appendix for the study of influential factors of scale control.

% \WW{I'd not say quantitative evaluation for scale control is not conducted here, just say we have additional results in appendix}.

\begin{figure}[!tb]
\setlength{\belowcaptionskip}{-0.6cm}
\centering
\includegraphics[width=1\linewidth]{  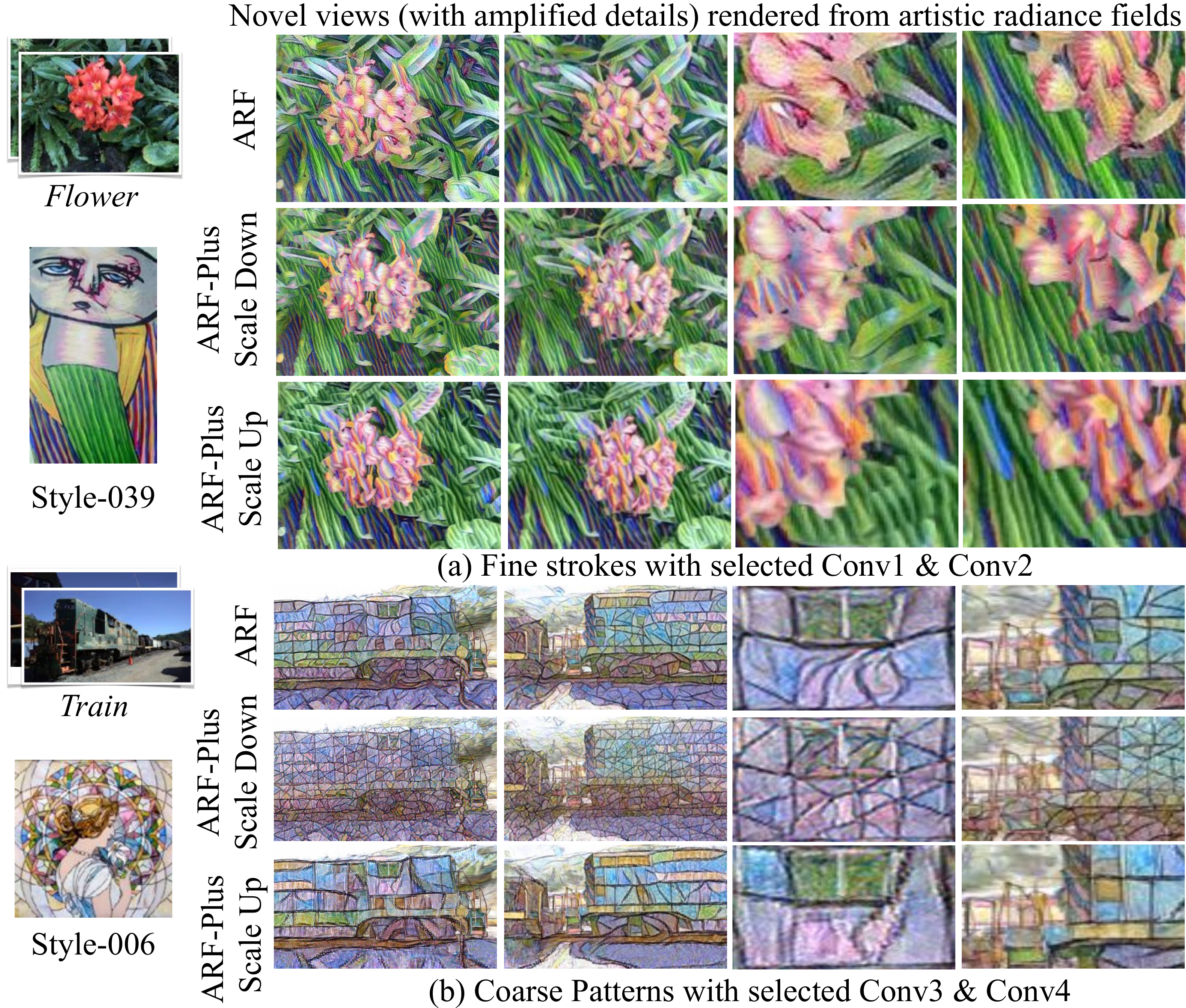}
\caption{Qualitative results of single style scale control. Our scale control effectively achieves style scaling. Please refer to supplementary materials for better visualization. }
\label{fig:scale_control_both_new}
\end{figure}

\subsubsection{Single Style Scale Control} 
% \noindent \textbf{Single Style Scale Control}\quad 
Fig. \ref{fig:scale_control_both_new} illustrates the successful scaling up and scaling down of style patterns in ARF-Plus. In addition, this approach provides enhanced flexibility for resizing fine strokes or coarse patterns by selectively assigning different weights to various receptive fields and generating diverse resized style images for each receptive field. In Fig. \ref{fig:scale_control_both_new} (a), by configuring the range of receptive field, specifically selecting Conv1 and Conv2 for feature extraction (setting $w_{l_1}=0.3, w_{l_2}=0.7$ and the rest to 0 in Eq. \ref{eq:scale control single loss style}), style transfer places greater emphasis on capturing fine details from the style image. It can be observed that the fine strokes - the green and white lines - are successfully scaled up or down in the stylized scenes. In Fig.  \ref{fig:scale_control_both_new} (b), by selecting Conv3 and Conv4 (setting $w_{l_3}=0.7, w_{l_4}=0.3$ and the rest to 0 to in Eq. \ref{eq:scale control single loss style}), style transfer focuses more on learning coarse patterns. Different layers are then assigned different values of the resize setting of the style image to achieve a continuously adjustable scale. The results show that the coarse shape patterns are effectively scaled up \/ down.

% \subsubsection{Multiple Style Scale Control}

\subsubsection{Multiple Style Scale Control}

\begin{figure}[!t]
\setlength{\belowcaptionskip}{-0.4cm}
\centering
\includegraphics[width=1\linewidth]{  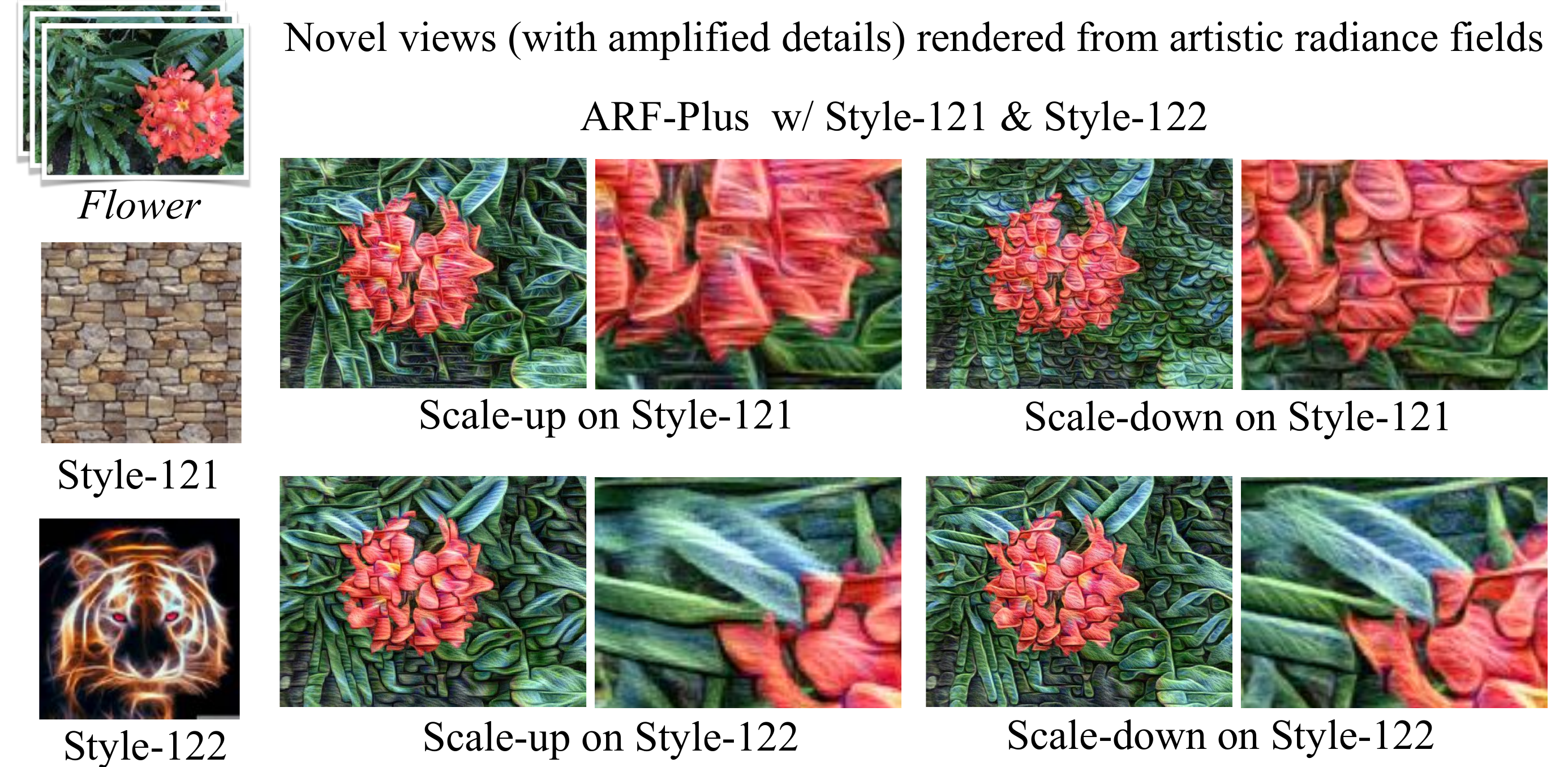}
  \caption{Qualitative results of multiple styles scale control. For better visualization, color is preserved by using our color preservation control. Our ARF-Plus with scale control can selectively scale up or scale down a particular style in stylization with blended multiple styles. }
  \label{fig:scale_control_styles_blend}
\end{figure}

% \noindent \textbf{Multiple Style Scale Control}\quad
After analyzing how well our scale control on individual styles, we move on to examine its effectiveness in handling multiple blended styles. We utilize color preservation to minimize the visual influence of different colors from multiple styles, enabling a more effective comparison of pattern scale variations. Fig. \ref{fig:scale_control_styles_blend} illustrates how our ARF-Plus scale control method, when applied to blended styles, effectively controls each individual style's scale. The noticeable changes in the size of leaf patterns are evident when scaling up or down the coarse pattern of Style-121. When the fine pattern scale changes for Style-122, there are observable effects on the leaves and petals.

\subsection{Spatial Control}

% \subsubsection{Single Style Spatial Control} 
\subsubsection{Single Style Spatial Control}
\begin{figure}[!t]
\setlength{\belowcaptionskip}{-0.4cm}
\centering
\includegraphics[width=1\linewidth]{   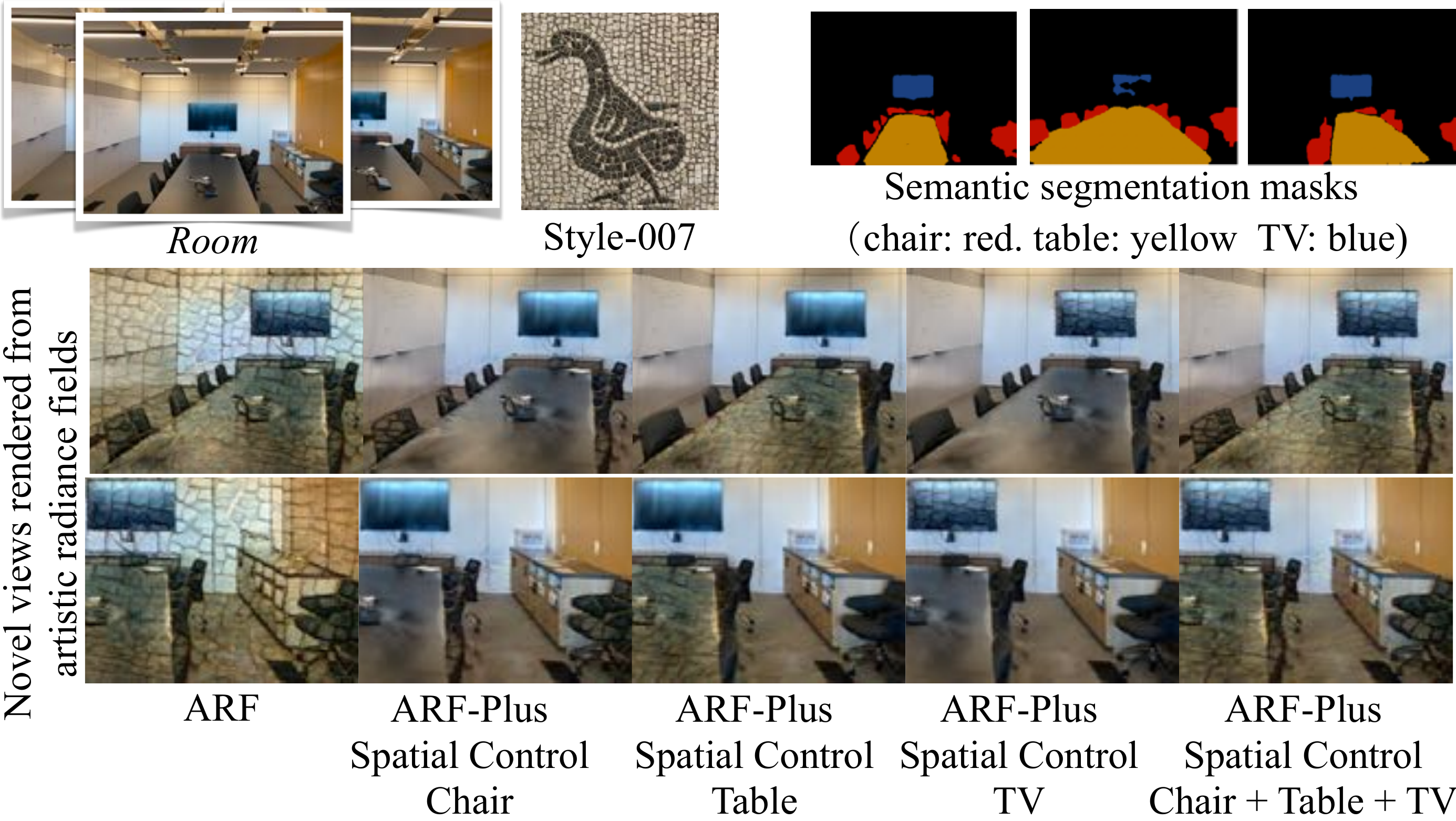}
  % \caption{Qualitative results of depth-aware control: style transfer on real-world forward-facing scenes - \textit{Leaves} and \textit{Flower}. Our method, ARF-Plus with Depth Control, demonstrates superior performance in preserving perceptive depth across varying views - the forefront leaf in (a), and the flower's petals in (b). Supplementary videos are provided [\href{https://drive.google.com/drive/folders/15zoJA8WtGAZs-JQGzIWGEDmcPFfPa_Xa?usp=sharing}{\textcolor{red}{click}}].}
  
  \caption{Qualitative results of spatial control with semantic segmentation masks: style transfer on the forward-facing scene - \textit{Room}. Our ARF-Plus with spatial control effectively stylizes specific semantic objects - chair, table, and TV - within the scene.}
  \label{fig:spatial_control_semantic_room}
\end{figure}

Fig. \ref{fig:spatial_control_semantic_room} presents the results of spatial control with semantic segmentation masks. It's worth noting that we use segmentation solely to produce masks $\mathbf{T}^{r}$. Our method emphasizes achieving spatial control given $\mathbf{T}^{r}$; the specific segmentation technique chosen isn't our primary focus, hence we haven't compared it with 3D segmentation methods. Our method successfully stylizes the associated elements of the scene \textit{Room}, namely the chairs, table, and TV. A noteworthy observation is that, despite the semantic segmentation masks being occasionally inaccurate or not fully covering the objects, our method manages to effectively paint nearly whole objects' areas in the stylized scene. One possible explanation is multiple training perspectives partially offset the shortcomings of semantic segmentation. Although certain views lead to poor segmentation results, alternative views' semantic maps provide broader and more accurate coverage. This enables the successful coloring of the overlapping regions across multiple views. Ultimately, the entire area of the semantic object is successfully stylized. 

\begin{table*}[!t]
% \begin{table}[h]
\footnotesize
\vspace{-5mm}
\setlength{\abovecaptionskip}{0pt}
\setlength{\belowcaptionskip}{0pt}
\caption{Quantitative results of spatial control with semantic segmentation masks. Evaluation metric ArtFID~\cite{wright2022artfid} is a combination of ContentDist and StyleFID. Room-All represents the combined usage of three semantic masks: Chair, Table, and TV. Results better than the baseline in the mask are in bold.
}
\smallskip
\renewcommand\arraystretch{0.5}
\centering
\setlength{\tabcolsep}{1mm}{
\resizebox{0.8\textwidth}{!}{
\begin{tabular}{ll|cc|cc|cc}
\toprule
% \smallskip
\multicolumn{2}{c}{} &
\multicolumn{2}{c}{ArtFID $\downarrow$ } & 
\multicolumn{2}{c}{ContentDist $\downarrow$} & 
\multicolumn{2}{c}{StyleFID $\downarrow$}\\
\midrule
 Style & Scene & ARF & ARF-Plus  w/ & ARF &  ARF-Plus  w/ &  ARF & ARF-Plus  w/ 
 \\
 & &  &  {\footnotesize Spatial Control } &  & {\footnotesize Spatial Control} &  & {\footnotesize Spatial Control} 
 \\
   & &  {\footnotesize in Mask} & {\footnotesize in Mask}  & in Mask & in Mask &  in Mask & in Mask \\
  % &  &  & Hist.match & Lumin.only & & Hist.match  & Lumin.only  && Hist.match  & Lumin.only  \\
 \midrule
% \multirow{2}{*}{
   
% \begin{minipage}{.09\textwidth}
%     Style-000 \\
%     \includegraphics[height=14mm, left]{figure/style_image_000.png}
% \end{minipage}

% } 

% & Leaves & 58.8668 & \textbf{56.3302} & \textbf{51.8628} & 1.6102 & \textbf{1.4568} & \textbf{1.4138} & 36.5598 & 38.6682 & 36.6845 \\

% & Flower & 65.977 & \textbf{65.0435} & \textbf{60.1018} & 1.6264 &\textbf{1.5526} & \textbf{1.4823} & 40.5667 & 41.2118 & \textbf{40.5472} \\

% & Trex & 67.031 & \textbf{61.8098} & \textbf{62.3082} & 1.6736 & \textbf{1.4998} & \textbf{1.4869} & 40.0509 & 41.2118 & 41.9049 \\

% & Horns & 64.0072 & \textbf{55.8469} & \textbf{55.9957} & 1.6083 & \textbf{1.5224} & \textbf{1.4816} & 39.7976 & \textbf{36.6831} & \textbf{37.7933} \\

% \midrule
007 & \textit{Room}-Chair  & 41.4614 & \textbf{39.3775}  & 0.0435 & \textbf{0.0434} & 38.7340 & \textbf{36.7384} \\

007 & \textit{Room}-Table  & 43.2753 & \textbf{40.8616}  &  \textbf{0.0473} & 0.0489 & 40.3191 & \textbf{37.9581} \\

007 & \textit{Room}-TV  & 41.2685 & \textbf{38.4802}  & 0.0195 & \textbf{0.0193} & 39.4799 & \textbf{36.7500} \\

007 & \textit{Room}-All  & 41.0287 & \textbf{37.9573} & \textbf{0.0950} & 0.0983 & 36.4707 & \textbf{33.5591} \\

% \midrule
% \multirow{2}{*}{

% \begin{minipage}{.09\textwidth}
%    % \smallskip
%     Style-007 \\
%     \includegraphics[height=14mm, left]{figure/style_image_007.png}
% \end{minipage}

% } 

% 122 & \textit{Horse}-horse &  47.8123 &  32.0022 & 33.9371 &  0.404 &  0.0917 &  0.1028 &  33.0536 &  28.3152 & 29.7731 \\

% 89 & \textit{Horse}-horse & 33.2293 &  23.1873  & \textbf{21.8684} & 0.4061  &  0.0902 & 0.1003 & 22.633 & 20.2693 &  \textbf{18.8756} \\

% 100 & \textit{Horse}-horse & 43.1613 &  31.215 & \textbf{30.1206} &  0.5085 &  0.1084 & 0.112 &  27.6116 & 27.1618 & \textbf{26.0861}\\

% & Horns & 38.3892 & 59.3562 & 44.4767& 1.5088 &  \textbf{1.4562} & \textbf{1.4554} & 25.4442 & 40.7622 & 30.5600 \\

\bottomrule
\end{tabular}}}
\label{table:spatial-control-semantic-mask}
\vspace{-5mm}
\end{table*}

We set the number of rendered novel views $N$ to 120 and render a series of images from the stylized radiance fields.
Table \ref{table:spatial-control-semantic-mask} provides quantitative results of the stylized scenes. In comparison to ARF in the mask, our spatial control yields overall superior ArtFID scores~\cite{wright2022artfid}, which can be attributed to a significant enhancement in styleFID. There are potentially two reasons behind the improvement in masked styleFID scores. Firstly, spatial control leads to a reduction in the regions requiring stylization within the scene. This results in fewer areas participating in gradient back-propagation during the same training epoch, allowing for the acquisition of more style information. Secondly, the appearance of semantic objects (such as TV) often exhibits similar textures. The uniform texture makes it easier for those areas to learn the style. Please refer to the Appendix for the explanation of calculating ArtFID on specific areas and additional qualitative evaluation. It is also notable that in Scene \textit{Room}, for ARF-Plus, the individual use of three separate semantic masks results in higher StyleFID values compared to the combined use of all three masks as a single mask (Room-All). Because ArtFID integrates both ContentDist and StyleFID, changes in StyleFID contribute to an elevated ArtFID score. This reveals that the size of the area controlled by spatial control has an impact on the effectiveness of style transfer. When the mask is too small, indicating a limited stylized area, it becomes difficult for that region to encompass all the style patterns depicted in the style image. Please refer to the Appendix for the explanation of calculating ArtFID on specific areas and additional qualitative evaluation.

\subsubsection{Multiple Style Spatial Control} 
\begin{figure}[!h]
\setlength{\belowcaptionskip}{-0.4cm}
\centering
\includegraphics[width=1\linewidth]{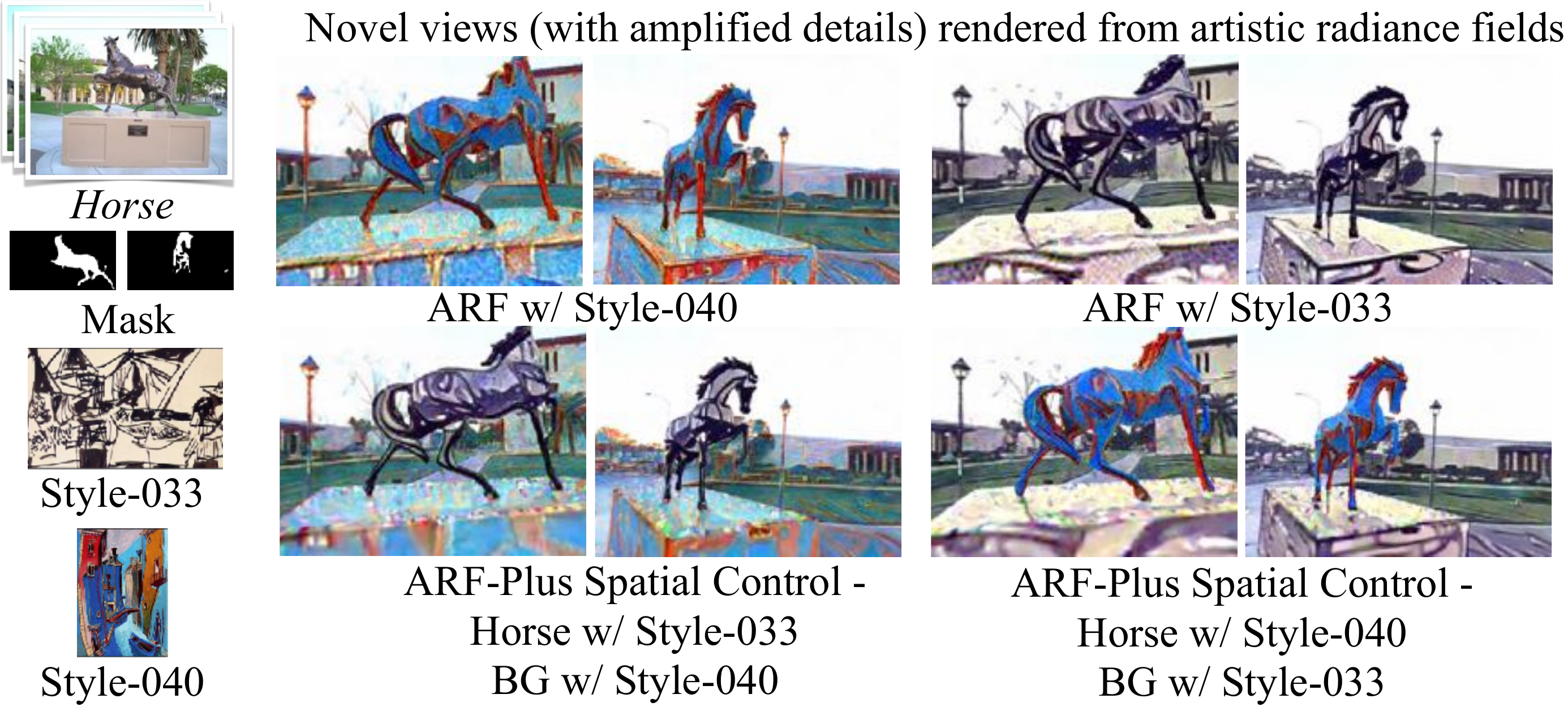}
  \caption{Qualitative results of multiple styles spatial control. Our multi-style spatial control successfully stylizes various locations with distinct styles. Please refer to supplementary materials for better visualization.}
  \label{fig:spatial_control_blend_styles}
\end{figure}

As shown in Fig. \ref{fig:spatial_control_blend_styles}, compared to the baseline's unified stylization, our proposed spatial control method effectively transfers various styles to different regions. The modification of the loss function (in Eq. \ref{eq: spatial-control-multi-loss-function}) may result in minor variations in the details. More specifically, our method calculates differences solely between a specific region and its corresponding style and segmented photo-realistic view. In contrast, the baseline ARF~\cite{zhang2022arf} computes the loss differences between the entire scene, style, and the overall photo-realistic view. Please refer to the Appendix for intermediate training results.

\subsection{Depth-aware Control}
\begin{figure}[!h]
\setlength{\belowcaptionskip}{-0.4cm}
\centering
\includegraphics[width=1\linewidth]{   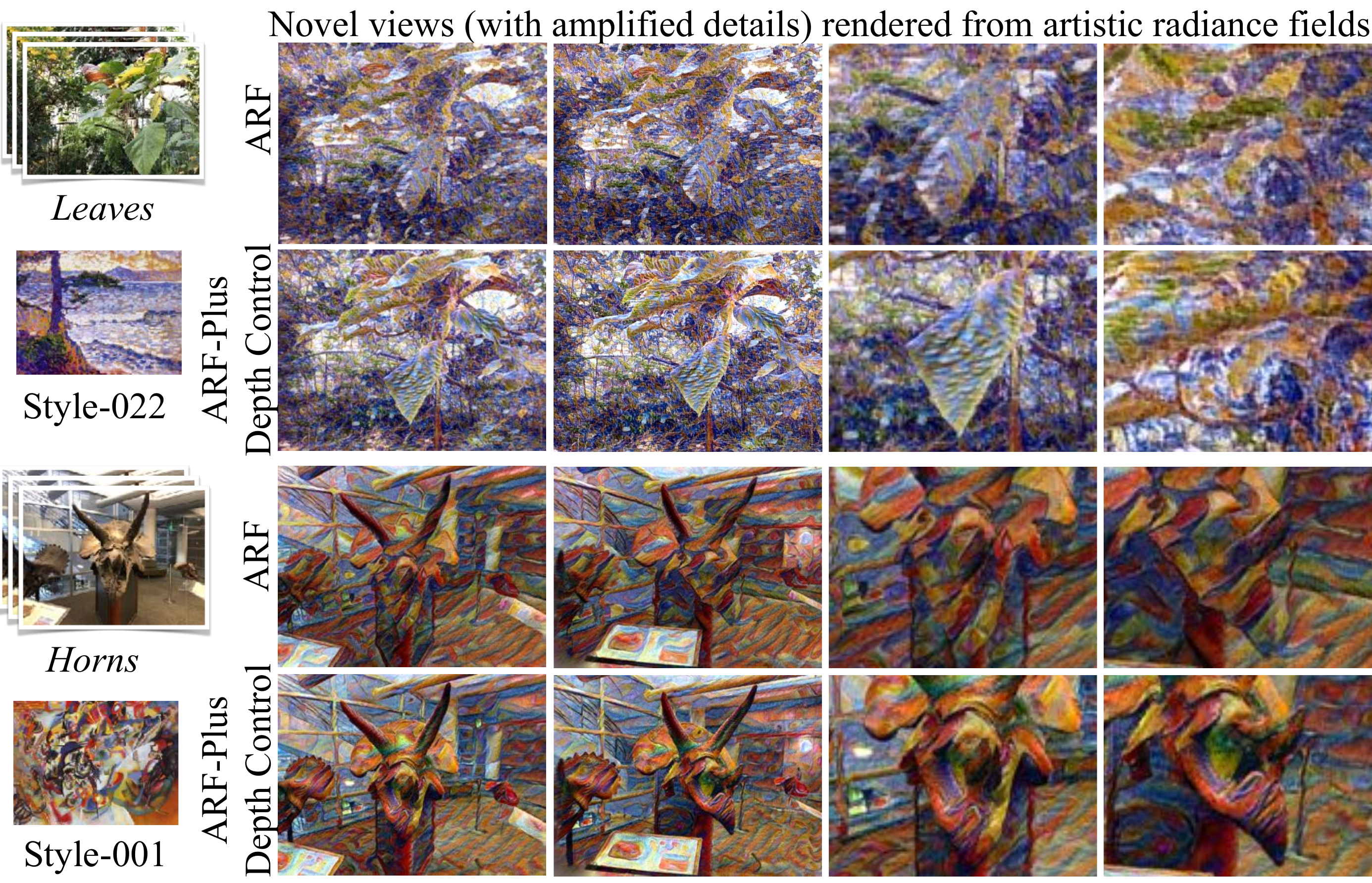}
  \caption{Qualitative results of depth enhancement control. Our depth control demonstrates superior performance in preserving perceptive depth across varying views - the forefront leaf in \textit{Leaves}, and the nostril (hole) in \textit{Horns}. Please refer to supplementary materials for better visualization.}
\label{fig:depth_control_face_forward}
\end{figure}

Fig. \ref{fig:depth_control_face_forward} presents comparisons between our depth-aware control method and the baseline. Qualitative results demonstrate that our method can successfully preserve perceptive depth across varying views. The foreground in \textit{Leaf} has a more distinct outline and stands out from the other leaves. For the \textit{Horns}, we better preserve the nostril (hole) located next to the horn. Please refer to the Appendix for quantitative evaluations.

\subsection{User Study}
We introduce user subjective evaluation to strengthen our results. Participants are shown a series of stylization results, each accompanied by the given style image(s), and stylized videos - one generated using ARF-Plus and the others using a baseline method with each single style image. They are asked to rate the ARF-Plus results of style transfer under specific controls. Ratings range from 0 (no discernible control) to 2 (highly noticeable control effect), with 1 indicating acceptability but occasional lack of prominence in certain areas or angles. Overall, we gather ratings for 28 randomly selected (scene, style(s)) pairs from various controls. 30 users were tasked with rating a randomly selected batch. We obtained approximately 15 ratings for each individual pair. We measure user satisfaction by dividing the sum of users' ratings by the total possible maximum score (no. of ratings * 2). The satisfaction results we obtained: 100 \% for Color Control (4 pairs), 90 \% for Single Style Scale Control (4 pairs), 83.33 \% for Multiple Style Scale Control (4 pairs), 96.67 \% for Single Style Spatial Control (4 pairs), 93.33 \% Multiple Style Spatial Control (4 pairs), 80 \% for Depth Control (4 pairs) and 93.33 \% for Combination of Controls (4 pairs). These results demonstrate the effectiveness of our methods.

\subsection{Combination of Controls}

% Figure \ref{fig:combination_controls}  demonstrates the combined utilization of our various controls. Our combination effect seamlessly integrates two styles into the same 3D scene, yielding flexible perceptual controls that the ARF baseline is unable to achieve. 

\begin{figure}[t]
\centering
\setlength{\belowcaptionskip}{-0.4cm}
\includegraphics[width=1\linewidth]{  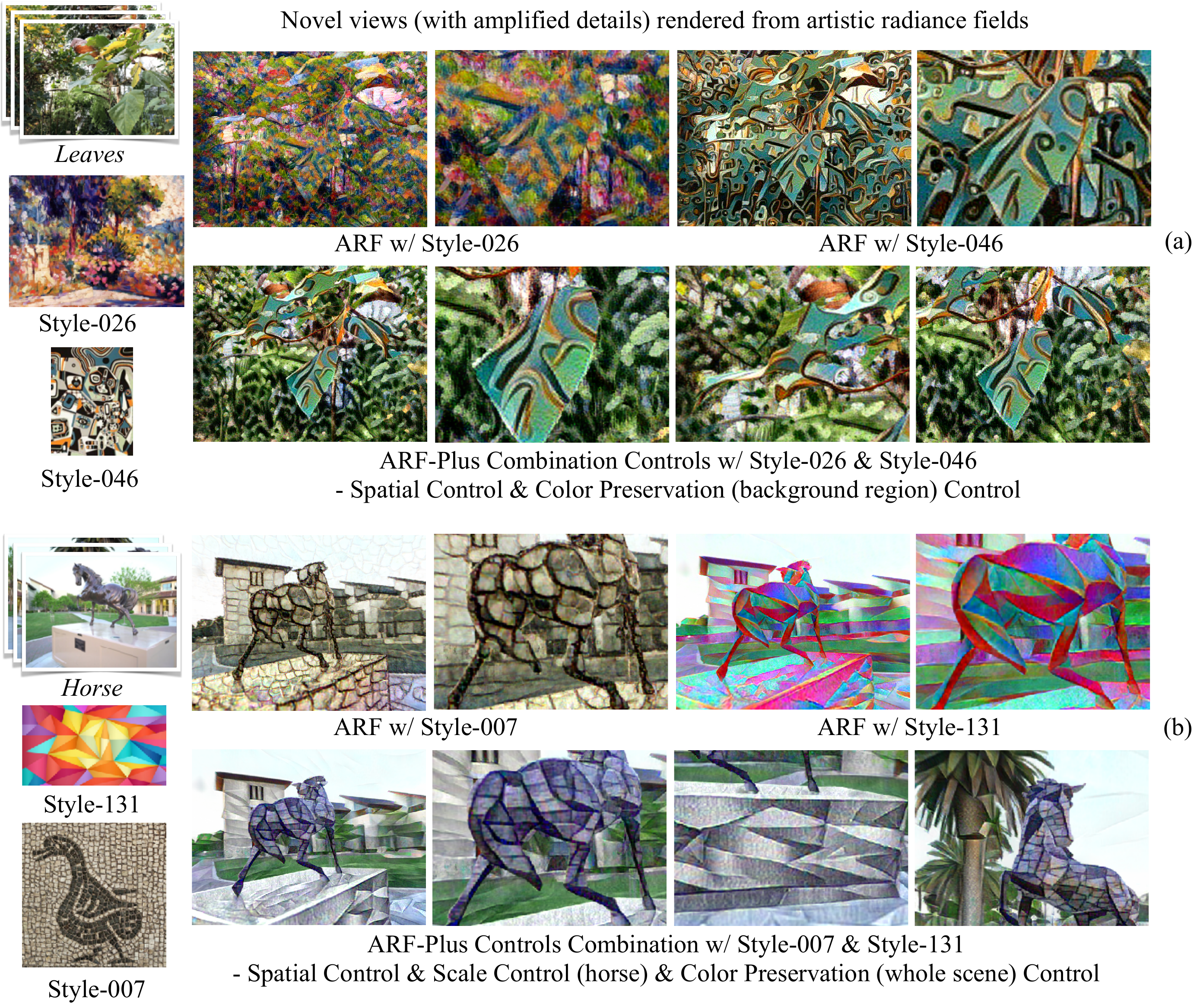}
  \caption{Qualitative results of controls combination. Please refer to supplementary materials for better visualization.}
  \label{fig:combination_controls}
\end{figure}

% In Figure \ref{fig:combination_controls} (a), through spatial control, we apply the observable pattern from style-046 to the primary leaf, while giving the background a pointillism impression. Additionally, the color-preservation option prevents style-026's extremely fluctuating colors from entering the scene. 
The previous experiments and evaluations demonstrated the effectiveness of our proposed control in four aspects: color, scale, spatial, and depth. These independent controls are an advantage because they allow for disentangled manipulation of each control individually. The following experimental results show our proposed controls can also be combined effectively, which enhances the overall controllability and flexibility of the framework. In Figure \ref{fig:combination_controls} (a), through spatial control, we apply the observable pattern from Style-046 to the primary leaf, while giving the background a pointillism impression as Style-026. Additionally, the colour-preservation option prevents Style-026's extremely fluctuating colours from entering the scene. In Figure \ref{fig:combination_controls} (b), the 3D scene has successfully been infused with distinct shapes from two different styles. Through spatial control, the semantic object (horse) has been adorned with regular cuboid shapes transferred by Style-007, while the rest of the background showcases triangular patterns from Style-131. It avoids monotony by incorporating multiple shapes from different styles. This imbues the entire scene with an abstract sensation, particularly noticeable in elements like the leaves of trees. Furthermore, in the ARF baseline, the shape of the horse sculpture appeared disproportionately large, giving the impression of significant fractures in the sculpture. With our scale control, the patterns on the horse sculpture have been reduced to a more suitable size, enhancing visual aesthetics. Moreover, the colour preservation feature allows us to freely select style images without excessive concern for whether their colours match our scene. In conclusion, compared to ARF, our control methods provide greater flexibility for 3D scene stylization, opening up endless possibilities for the imagination.

\section{Summary}
In this paper, we highlight a research gap in radiance fields style transfer and emphasize the significance of various perceptual controls in 3D stylization, catering to diverse user needs across various application scenarios. Our approach offers nontrivial modifications and novel strategies unique to 3D models, distinct from direct adaptations of 2D stylization techniques. For color control, we introduce a novel unified loss function with a luminance channel for the radiance fields model, excluding the "channel-concatenation" approach suited only for 2D image style transfer. For scale control, a novel formula is introduced to enhance pattern variety and smoothness. In spatial control, we utilize masks on cached-gradient maps (deferred back-propagation strategy) to deal with the losses in various angles for 3D radiance field representation. Additionally, we pioneer depth control for 3D stylization, and propose flexible methods for multiple style control in scale and spatial aspects. Both quantitative and qualitative evaluations on real-world datasets demonstrate that our ARF-Plus effectively achieves perceptual control for 3D scene stylization. All types of our proposed style controls are implemented in an integrated ARF-Plus framework, which can be transformed into a user-friendly app, allowing arbitrary combinations of different style controls for boundless creative freedom. We anticipate our research will inspire further exploration in this innovative field.

\clearpage

%%%%%%%%% REFERENCES
{\small
\bibliographystyle{wacv}
\bibliography{wacv}
}

\end{document}